\documentclass{article}
\usepackage{cite}
\usepackage{amsmath,amssymb,amsfonts,amsthm}
\usepackage[ruled,vlined]{algorithm2e}
\usepackage{enumerate}
\usepackage{graphicx}
\usepackage{textcomp}
\usepackage{psfrag}
\usepackage{url}
\usepackage[all]{xy}
\usepackage{multirow}
\def\BibTeX{{\rm B\kern-.05em{\sc i\kern-.025em b}\kern-.08em
    T\kern-.1667em\lower.7ex\hbox{E}\kern-.125emX}}

\newcommand{\R}{\mathbb{R}}

\newcommand{\Q}{\mathbb{H}}

\newcommand{\A}{\mathbb{A}}

\newcommand{\K}{\mathbb{K}_4}

\newcommand{\vetx}{\boldsymbol{x}}
\newcommand{\vety}{\boldsymbol{y}}

\newcommand{\vett}{\boldsymbol{t}}

\newcommand{\matA}{\boldsymbol{A}}
\newcommand{\matB}{\boldsymbol{B}}
\newcommand{\matC}{\boldsymbol{C}}

\newcommand{\matX}{\boldsymbol{X}}

\newcommand{\matW}{\boldsymbol{W}}
\newcommand{\matM}{\boldsymbol{M}}
\newcommand{\matT}{\boldsymbol{T}}
\newcommand{\matH}{\boldsymbol{H}}

\newcommand{\bb}{\begin{equation}}
\newcommand{\ee}{\end{equation}}
\newcommand{\bbb}{\begin{eqnarray}}
\newcommand{\eee}{\end{eqnarray}}
\newcommand{\benu}{\begin{enumerate}}
\newcommand{\eenu}{\end{enumerate}}

\newcommand{\bpm}{\begin{bmatrix}}
\newcommand{\epm}{\end{bmatrix}}

\newcommand{\ii}{\boldsymbol{i}}
\newcommand{\jj}{\boldsymbol{j}}
\newcommand{\kk}{\boldsymbol{k}}

\newcommand{\hyperN}[1]{{#1}_0 + {#1}_1 \boldsymbol{i}_{1} +  \dots + {#1}_n \boldsymbol{i}_{n}}

\def\boxmax{\kern 0em\hbox{\rm \kern .25em\lower.1ex\hbox{\rlap{$\vee$}}\kern -.07em\lower.2ex\hbox{$\square$}\kern.25em}}
\def\boxmin{\kern 0em\hbox{\rm \kern .25em\lower.1ex\hbox{\rlap{$\wedge$}}\kern -.07em\lower.2ex\hbox{$\square$}\kern.25em}}
\def\boxdiamond{\kern 0em\hbox{\rm \kern .25em\hbox{\rlap{$\diamond$}}\kern -.15em\lower.2ex\hbox{$\square$}}}

\newtheorem{definition}{Definition}

\theoremstyle{definition}
\newtheorem{example}{Example}
\newtheorem{remark}{Remark}

\begin{document}

\title{A General Framework for Hypercomplex-valued Extreme Learning Machines
\thanks{This work was supported in part by CNPq under grant no. 310118/2017-4, FAPESP under grant no. 2019/02278-2, and Coordena\c{c}\~ao  de Aperfei\c{c}oamento de Pessoal de N\'ivel Superior - Brasil (CAPES) - Finance Code 001.}
}

\author{Guilherme Vieira and Marcos Eduardo Valle
\thanks{Guilherme Vieira and Marcos Eduardo Valle are with the Department of Applied Mathematics, University of Campinas, Campinas -- SP, Brazil. Emails: vieira.g@ime.unicamp.com.br and valle@ime.unicamp.br.}
}

\maketitle

\begin{abstract}
This paper aims to establish a framework for extreme learning machines (ELMs) on general hypercomplex algebras. Hypercomplex neural networks are machine learning models that feature higher-dimension numbers as parameters, inputs, and outputs. Firstly, we review broad hypercomplex algebras and show a framework to operate in these algebras through real-valued linear algebra operations in a robust manner. We proceed to explore a handful of well-known four-dimensional examples. Then, we propose the hypercomplex-valued ELMs and derive their learning using a hypercomplex-valued least-squares problem. Finally, we compare real and hypercomplex-valued ELM models' performance in an experiment on time-series prediction and another on color image auto-encoding. The computational experiments highlight the excellent performance of hypercomplex-valued ELMs to treat high-dimensional data, including models based on unusual hypercomplex algebras. 
\end{abstract}


\section{Introduction}

Hypercomplex algebras over the real field are crucial for modern mathematics, physics, and applied areas such as computer graphics and computational intelligence. For example, complex numbers are vital for digital signal processing, and information theory \cite{oppenheim89}. Quaternions are of unparalleled value for modeling 3-dimensional movements, such as in graphic design and automated control \cite{kuipers99}. The Klein four-group is deeply connected to graph theory \cite{jose15} and was for the construction of hypercomplex-valued Hopfield neural networks \cite{Kobayashi2020HopfieldFour-group}. The tessarines form a commutative hypercomplex number system successfully used for signal processing \cite{alfsmann06}. Clifford algebras, a family of algebras with the power of two dimensions, pose a reliable generalization of geometric algebra and have an essential role in digital image processing \cite{labunets04}. This paper discusses hypercomplex algebras' critical concepts to develop a neural network model for multi-dimensional data processing.

Hypercomplex-valued neural networks (HvNNs) are defined as extensions of traditional real-valued neural network models to hypercomplex number systems. In a few words, HvNNs possess a similar architecture to real-valued model they derive from but the adjustable parameters as well as their inputs and outputs are elements of a hypercomplex algebra. Known examples of HvNNs are the complex-valued \cite{aizenberg11book,hirose12}, hyperbolic-valued \cite{buchholz00,nitta08,nitta18}, quaternion-valued \cite{xia18,parcollet19air}, and octonion-valued neural networks \cite{popa16b,castro17cnmac}. These architectures are well adapted to multi-signal processing, meaning they can cope appropriately with phase information and rotations, for instance. Applications of HvNNs mainly revolve around signal and image processing \cite{minemoto16,xia15,xu16,castro17bracis,chen17,papa17,xiaodong17,kinugawa18,aizenberg18wcci,wang19}, classification and prediction \cite{Ujang2011Quaternion-valuedFiltering,shang14,talebi15,popa16a,minemoto17,greenblatt18}, and general image treatment such as auto-encoding and denoising \cite{minemoto17,Vieira2020ExtremeAuto-Encoding}.

Extreme learning machines (ELM) are a well-established type of feedforward neural networks introduced by Huang \cite{huang04} in the early 2000s. This simple architecture consists of a fully connected multilayer feedforward network in which all but the output layer have fixated randomly initialized parameters. Furthermore, a least-squares optimization problem performs training on the last layer. The ELMs maintain the universal approximation capabilities of a multilayer Perceptron while also drastically decreasing training's computational complexity \cite{huang06,huang11a,wang11}. Complex-valued and quaternion-valued ELMs have been developed respectively by Li et al. \cite{li05}, Minemoto et al. \cite{minemoto17}, and Lv et al. \cite{lv18}. This paper extends the ELMs to more general hypercomplex number systems. Specifically, this work extends a conference paper where we observed that ELMs based on Cayley-Dickson algebras outperformed real-valued and quaternion-valued ELMs with a comparable number of trainable parameters in an auto-encoding task \cite{Vieira2020ExtremeAuto-Encoding}. Our motivation is to further investigate ELMs on general hypercomplex algebras as tools for high-dimensional data processing, common to image processing, time series forecasting, and general classification and regression tasks. Concluding, in this paper, we define key concepts of hypercomplex-valued extreme learning machines. Moreover, we address applications on times series prediction and auto-encoding tasks. 

The paper is organized as follows: Section \ref{sec:hypercomplex} presents the basic concepts on hypercomplex numbers. Hypercomplex matrix operations and their equivalence to real-valued linear algebra are detailed in Section \ref{sec:hc-matrix-algebra}. Section \ref{sec:CD-ELM} introduces the hypercomplex ELMs along with their training algorithm. Computational experiments featuring four-dimensional algebras for times series prediction and color image auto-encoding are described in Section \ref{sec:Applications}. The paper finishes with the concluding remarks in Section \ref{sec:concluding-remarks}.

\section{A Brief Review of Hypercomplex Algebras} \label{sec:hypercomplex}

Let us begin by recalling the core concepts of hypercomplex number systems. Although hypercomplex algebras are usually defined over an arbitrary field $\mathbb{F}$, we only consider real numbers as the ground field in this paper.

A hypercomplex number $x$ has a representation in the form
\bb \label{eq:hc-number} x = \hyperN{x}, \ee
where $x_0,x_1,\dots,x_n \in \mathbb{R}$. The elements $\boldsymbol{i}_1, \boldsymbol{i}_2,\dots,\boldsymbol{i}_n$ are called hyperimaginary units \cite{kantor89,castro20nn}. We denote the set of all hypercomplex numbers by $\mathbb{A}$.

A hypercomplex algebra is obtained by enriching the set of hypercomplex numbers $\mathbb{A}$ with an addition and a multiplication \cite{kantor89,castro20nn}. The addition is performed in a component-wise manner by means of the equation
\begin{equation} \label{eq:hc-addition}
    x+y = (x_0+y_0)+(x_1+y_1)\boldsymbol{i}_1+\ldots+(x_n+y_n)\boldsymbol{i}_n,
\end{equation}
for hypercomplex numbers $x=\hyperN{x}$ and $y=\hyperN{y}$. 

The multiplication is performed distributively using the interactions between imaginary units. Precisely, we first define the product between any two hyperimaginary unit 
\begin{equation} \label{eq:mu_ij} 
\ii_i \ii_j \equiv \mu_{ij} = \hyperN{(\mu_{ij})} \in \mathbb{A},
\end{equation} 
for all $i,j=1,\ldots,n$. The hypercomplex numbers $\mu_{ij}$, which are usually provided by a multiplication table, characterize the hypercomplex algebra. The multiplication of two hypercomplex numbers $x = \hyperN{x}$ and $y = \hyperN{y}$ is defined using the distributive law and replacing the product $(x_i\ii_i)(y_j \ii_j)$ by $x_i y_j \mu_{ij}$, for all $i,j=1,\ldots,n$. Formally, we have
\begin{equation}
\begin{aligned} \label{eq:multiplication}
xy &= \left(x_0 y_0 + \sum_{i,j=1}^n x_i y_j (\mu_{ij})_0 \right) \\  & 
+ \left(x_0y_1 + x_1 y_0+  
\sum_{i,j=1}^n x_i y_j (\mu_{ij})_1 \right)\ii_1 + \ldots \\ &+ \left( x_0 y_n + x_ny_0 + \sum_{i,j=1}^n x_i y_j (\mu_{ij})_n \right)\ii_n.
\end{aligned}
\end{equation}

Unlike in a field, the multiplication of hypercomplex numbers need not be associative, commutative, nor have many other algebraic properties. 
Nevertheless, a scalar $\alpha \in \R$ can be identified with the hypercomplex number $\alpha+0\ii_1+\ldots+0\ii_n$. Moreover, the scalar product 
\begin{equation} \label{eq:scalar_product}
\alpha x = \hyperN{\alpha x},
\end{equation} 
can be derived from \eqref{eq:multiplication}.
%
% Concluding, an hypercomplex algebra is defined as follows \cite{kantor89}:
% \begin{definition}[Hypercomplex algebra]
% A hypercomplex algebra $\A$ of dimension $n$ over the real numbers $\mathbb{R}$ is a set of hypercomplex numbers of the form $$p = p_1 \e{1} + p_2 \e{2} + \dots + p_n \e{n}; \quad p_1,\dots,p_n \in \mathbb{R}$$ equipped with addition and scalar product operations defined entry-wise as follows:
% \begin{itemize}
%     \item Addition: \bb  \label{eq:hc-addition} p+q = (p_1+q_1) \e{1} + \dots + (p_n+q_n) \e{n}.\ee 
%     \item Scalar product: \bb \label{eq:hc-scalar_product} \alpha p = \alpha p_1 \e{1} + \dots + \alpha p_n \e{n}.\ee 
% \end{itemize}
% \end{definition}
%
This leads to the existence of a canonical isomorphism between a hypercomplex algebra and the vector space $\mathbb{R}^{n+1}$. Precisely, the isomorphism $\varphi: \A \to \mathbb{R}^{n+1}$ is defined by
\begin{equation} \label{eq:isomorphism}
\varphi(x) = \bpm x_0 \\ x_1 \\ \vdots \\ x_n \epm, \quad \forall x=\hyperN{x} \in \mathbb{A}. 
\end{equation}
Clearly, $\varphi$ is linear because it consists of a simple rearrangement of the components of $x$. The inverse isomorphism $\varphi^{-1}$ is merely the inverse rearrangement operation. 

The isomorphism between an hypercomplex algebra $\mathbb{A}$ and the vector space $\mathbb{R}^{n+1}$ allows us to express the multiplication of hypercomplex numbers using a matrix-vector product. Formally, given a hypercomplex number $a \in \A$, the multiplication to the left by $a$ yields an operator $\mathcal{A}_L: \A \to \A$ defined by $\mathcal{A}_L (x) = ax$, for all $x\in\A$. It is not hard to verify that $\mathcal{A}:\mathbb{A} \to \mathbb{A}$ is a linear operator. As a consequence, using the isomorphism $\varphi$, we conclude that
\bb \label{eq:A_L-isomorphism} \varphi \left(\mathcal{A}_L(x) \right) = \Phi_L(a) \varphi(x),\ee 
where $\Phi_L:\A \to \R^{(n+1) \times (n+1)}$ is the matrix representation of $\mathcal{A}_L$ with respect to the canonical basis \cite{catoni05}, that is,
\bb \Phi_L(a) = \begin{bmatrix}
| & | & & | \\
\varphi(a) & \varphi(a\ii_1) & \ldots & \varphi(a\ii_n) \\
| & | & & | 
\end{bmatrix}. \ee
Note that the terms $a\ii_i$ are products in $\A$. Therefore, the operator $\Phi_L$ depends ultimately on the multiplication table of the algebra $\A$. In fact, the identity
\[ \Phi_L(a) = \begin{bmatrix}
a_0 & \sum_{i} a_i (\mu_{ij})_0 & \ldots & \sum_{i} a_i (\mu_{ij})_0 \\ 
a_1 & a_0 + \sum_{i} a_i (\mu_{ij})_1 & \ldots & \sum_{i} a_i (\mu_{ij})_1 \\ 
\vdots & \vdots & \ddots & \vdots \\ 
a_n & \sum_{i} a_i (\mu_{ij})_n & \ldots & \sum_{i} a_0 + a_i (\mu_{ij})_n 
\end{bmatrix}, \]
holds for all $a=\hyperN{a}$.

In a similar fashion, the multiplication to the right by $a \in \mathbb{A}$ yields an operator $\mathcal{A}_R: \A \to \A$ defined by $\mathcal{A}_R (x) = xa$, for all $x\in\A$. Using the isomorphism $\varphi$ given by \eqref{eq:isomorphism}, we obtain
\bb \label{eq:A_R-isomorphism} 
\varphi \left(\mathcal{A}_R(x) \right) =  \Phi_R(a) \varphi(x),\ee 
where $\Phi_R:\A \to \R^{(n+1) \times (n+1)}$ is defined by
\bb \Phi_R(a) = \begin{bmatrix}
| & | & & | \\
\varphi(a) & \varphi(\ii_1 a) & \ldots & \varphi(\ii_n a) \\
| & | & & | 
\end{bmatrix}. \ee

Finally, we define the absolute value (or norm) of a hypercomplex number $x=\hyperN{x} \in \A$ by means of the following equation:
\begin{equation}
    \label{eq:absolute-value}
    |x| = \sqrt{\sum_{i=0}^{n} |x_{i}|^2}.
\end{equation}
In other words, the absolute value of $x$ corresponds to the Euclidean norm of $\varphi(x)$, that is, $|x| = ||\varphi(x)||_2$.

% The collection $\left\lbrace \mu_{ij} \right\rbrace$ can be arranged in a matrix referred to as multiplication table.
% \[ M =  [\mu_{ij}] \]
%  Moreover, an algebra with $(n-1)$ imaginary units is said to have dimension $n$.

\section{Hypercomplex-Valued Matrix Algebra} \label{sec:hc-matrix-algebra}

This section uses the well-known real matrix algebra to represent operations in any hypercomplex matrix algebra. Furthermore, we propose a method to solve hypercomplex least-squares problems through an equivalent real least-squares problem.

A hypercomplex matrix is a matrix whose entries are elements of a hypercomplex algebra $\A$. Such as real-valued linear algebra, the product of matrices $\matA \in \A^{M\times L}$ and $\matB \in \A^{L \times N}$ results in a new matrix $\matC \in \A^{M \times N}$ with entries defined by
\bb \label{eq:matrix-product} c_{ij} = \sum_{\ell=1}^L a_{i\ell} b_{\ell j}, \ee
where $i=1,\ldots,M$ and $j=1,\ldots,N$. Here, $a_{i\ell}$ and $b_{\ell j}$ are entries of the matrices $\matA$ and $\matB$, respectively.

In order to take advantage of fast scientific computing softwares, in practice, we compute matrix operations using real-valued linear algebra using the isomorphism \eqref{eq:isomorphism} and either \eqref{eq:A_L-isomorphism} or \eqref{eq:A_R-isomorphism}. Precisely, by applying the isomorphism $\varphi$ in both sides of \eqref{eq:matrix-product} and using \eqref{eq:A_L-isomorphism}, we obtain 
\begin{align*}
    \varphi(c_{ij}) &  
    = \sum_{\ell=1}^L \varphi\left( a_{i\ell} b_{\ell j} \right)  
    = \sum_{\ell=1}^L \Phi_L(a_{i\ell})\varphi(b_{\ell j}) \\ 
    &= \begin{bmatrix}
    \Phi_L(a_{i1}) & \Phi_L(a_{i2}) & \ldots & \Phi_L(a_{iL})
    \end{bmatrix}
    \begin{bmatrix}
     \varphi(b_{1j}) \\ \varphi(b_{2j}) \\ \vdots \\ \varphi(b_{Lj})
    \end{bmatrix}.
\end{align*} 
Equivalently, using real-valued matrix operations, we have
\bb \label{eq:Matrix-vec-prod} \varphi(\matC) = \Phi_L(\matA) \varphi(\matB), \ee 
where $\Phi_L$ and $\varphi$ are defined as follows for hypercomplex matrix arguments:
\bb \label{eq:Matrix-Phi_L}
\Phi_L(\matA) = \begin{bmatrix}
\Phi_L(a_{11}) & \Phi_L(a_{12}) & \ldots & \Phi_L(a_{1L}) \\
\vdots  & \vdots & \ddots & \vdots \\
\Phi_L(a_{M1}) & \Phi_L(a_{M2}) & \ldots & \Phi_L(a_{ML}) 
\end{bmatrix},
\ee 
and
\bb \label{eq:Matrix-varphi}
\varphi(\matB) = \begin{bmatrix}
\varphi(b_{11}) & \ldots & \varphi(b_{1N}) \\
\varphi(b_{21}) & \ldots & \varphi(b_{2N}) \\
\vdots & \ddots & \vdots \\
\varphi(b_{L1}) & \ldots & \varphi(b_{LN}) \\
\end{bmatrix}.
\ee 
Note that $\Phi_L(\matA)$ is a real matrix of size $(n+1)M\times (n+1)L$ while $\varphi(\matB)$ is a real matrix of size $(n+1)L\times N$. The real-valued matrix $\varphi(\matC) \in \R^{(n+1)M\times N}$ is defined analogously to \eqref{eq:Matrix-varphi}. Furthermore, the hypercomplex matrix $\matC \in \A^{M\times N}$ can be obtained by rearranging the elements of $\varphi(\matC)$. Formally, reorganizing the elements of $\varphi(\matC)$ defines the inverse mapping $\varphi^{-1}: \R^{(n+1)M\times N} \to \A^{M \times N}$. More importantly, we have 
\bb \label{eq:CD2Real} 
\matC = \varphi^{-1} \left(\Phi_L(\matA) \varphi(\matB) \right), \ee 
which provides an effective formula for the computation of hypercomplex matrix product using the real-valued linear algebra often available in scientific computing softwares. 

Alternatively, it is possible to compute $\matC$ using the multiplication to the right by $b_{\ell j}$ in \eqref{eq:matrix-product} instead of multiplication to the left by $a_{i\ell}$. 
% Essentially, this means defining an operator analogous to \eqref{eq:CD2Real} for the multiplication to the right, $\mathcal{B}_R(x)=xb$ for all $x \in \A$. 
In this case, we have
\begin{align*}
    \varphi(c_{ij}) & = \sum_{\ell=1}^L  \Phi_R(b_{\ell j}) \varphi(a_{i\ell})  \\ 
    &= 
    \begin{bmatrix}
    \Phi_R(b_{1j}) & \Phi_R(b_{2j}) & \ldots & \Phi_R(b_{Lj})
    \end{bmatrix}
    \begin{bmatrix}
     \varphi(a_{i1}) \\ \varphi(a_{i2}) \\ \vdots \\ \varphi(a_{iL})
    \end{bmatrix},
\end{align*} 
for all $i=1,\ldots,M$ and $j=1,\ldots,N$. Using real-valued matrix operations, we obtain the identity
\bb \phi(\matC) = \Phi_R(\matB) \phi(\matA), \ee 
where $\Phi_R$ is defined by
\bb \label{eq:Matrix-Phi_R}
\Phi_R(\matB) = \begin{bmatrix}
\Phi_R(b_{11}) & \Phi_R(b_{21}) & \ldots & \Phi_R(b_{L1}) \\
\vdots  & \vdots & \ddots & \vdots \\
\Phi_R(b_{1N}) & \Phi_R(b_{2j}) & \ldots & \Phi_R(b_{LN})
\end{bmatrix},
\ee 
and $\phi(\matA) = \varphi(\matA^T)$. Note that $\Phi_R(\matB)$ and $\varphi(\matA^T)$ are  real-valued matrices of size $(n+1)N\times (n+1)L$ and $(n+1)L\times M$.
Hence, the hypercomplex-valued matrix $\matC = \matA \matB$ can be alternatively computed by means of the equation 
\bb \label{eq:CD2Real2} 
\matC = \phi^{-1} \left(\Phi_R(\matB) \varphi(\matA^T) \right). \ee 
\begin{remark}
From a computational standpoint, the bottleneck in \eqref{eq:CD2Real} and \eqref{eq:CD2Real2} is the construction of the real-valued matrices $\Phi_L(\matA)$ and $\Phi_R(\matB)$ of sizes $(n+1)M\times(n+1)L$ and $(n+1)N \times (n+1)L$, respectively. It can be seen that
\eqref{eq:CD2Real} is faster than \eqref{eq:CD2Real2} if the matrix $\matA$ has less entries than $\matB$, and vice-versa. We suggest implementing both \eqref{eq:CD2Real} and \eqref{eq:CD2Real2} and compute the product of two hypercomplex-valued matrices using the fastest formula.
\end{remark}

The training step of extreme learning machines (ELMs) is formulated as a least-squares problem. The framework described below can solve such least-squares problems in hypercomplex algebras, thus allowing the implementation of hypercomplex-valued ELMs. Let us begin the discussion on the hypercomplex-valued least-squares problem with the Frobenius norm.

Analogously to the real-valued case \cite{golub96}, the Frobenius norm of a hypercomplex-valued matrix $\matA \in \A^{M \times N}$ is defined by
\bb \label{eq:frob-hypercomplex} \|\matA\|_F = \sqrt{\sum_{i=1}^M \sum_{j=1}^N |a_{ij}|^2},\ee
where $|a_{ij}|$ represents the absolute value of $a_{ij} \in \A$. Combining \eqref{eq:isomorphism} and \eqref{eq:absolute-value}, we have $|a_{ij}| = \|\varphi(a_{ij})\|_2$, where $\|\cdot\|_2$ denotes the usual Euclidean norm. Thus, we have 
\bb \label{eq:norm-equivalence} \|\matA\|_F = \|\varphi(\matA)\|_F.\ee
Using the Frobenius norm \eqref{eq:frob-hypercomplex}, we define the hypercomplex least squares problem as follows:
\begin{definition}[Hypercomplex-Valued Least Squares Problem] \label{def:hypercomplex-LSP}
Given matrices $\matA \in \A^{M \times L}$ and $\matB \in \A^{M \times N}$, the hypercomplex-valued least squares problem consists of finding the minimal Frobenius norm solution to the problem
\bb \label{eq:hypercomplexLS} \min \left\{ \|\matA \matX - \matB\|_F: \matX \in \A^{L \times N} \right\}.\ee
\end{definition}

In practice we solve an hypercomplex-valued least square problem using the real-valued algebra framework detailed previously. From \eqref{eq:Matrix-vec-prod} and \eqref{eq:norm-equivalence}, we write
\begin{align*}
    \|\matA \matX - \matB\|_F 
    &= \| \varphi(\matA \matX - \matB)\|_F 
    = \| \varphi(\matA \matX) - \varphi(\matB)\|_F \\
    &= \| \Phi_L(\matA) \varphi(\matX) - \varphi(\matB)\|_F.
\end{align*}
Hence, the hypercomplex-valued least squares problem is rewritten as a real-valued problem:
\bb \label{eq:realLSP} \min\{\|\Phi_L(\matA)\matX^{(r)} - \varphi(\matB)\|_F: \matX^{(r)} \in \R^{(n+1)L\times N}\},\ee
where $\matX^{(r)}$ corresponds to $\varphi(\matX)$.
The real-valued least square problem given by \eqref{eq:realLSP} can be solved by means of the Moore-Penrose pseudoinverse \cite{golub96}, i.e.,
\bb \label{eq:solution-realLSP}
\matX^{(r)} = \Phi_L(\matA)^\dag \varphi(\matB),\ee 
where $\Phi_L(\matA)^\dag$ is the pseudoinverse of $\Phi_L(\matA)$. Concluding, making use of the real-valued linear algebra, the solution of the hypercomplex-valued least squares problem \eqref{eq:hypercomplexLS} is given by
\bb \label{eq:solution-hypercomplexLS} 
\matX = \varphi^{-1}\left(\Phi_L(\matA)^\dag \varphi(\matB) \right).\ee

\section{Hypercomplex-Valued Extreme Learning Machines} \label{sec:CD-ELM}

Extreme learning machines (ELMs) are feedforward neural network models in which all trainable parameters are located in the output layer. Specifically, the hidden layer parameters are randomly generated and fixated. Training the output layer parameters is formulated as a least-squares problem. Thus, training an ELM is achieved in a finite number of operations, allowing the model to achieve high-performance rates while also presenting very low computational cost \cite{huang04,huang06,huang11a,huang11b}. 

Let us define a single-hidden layer feedforward neural network on a hypercomplex algebra $\A$. The parameters of the single hidden layer with $L$ neurons are represented by a matrix $\matW \in \A^{D \times L}$. Given a  hypercomplex-valued row vector $\vetx = [x_1,\ldots,x_D] \in \A^D$ as input, the feed-forward step through the hidden layer yields
\bb \label{eq:hidden-layer} \boldsymbol{h} = f(\vetx \matW) \in \A^{L},\ee 
where $f:\A \to \A$ is a non-linear activation function defined in an entry-wise manner for matrices. The activation function $f$ is usually defined in a split manner, i.e., a real-valued non-linear function applied separately to each component of an hypercomplex argument. For example, the split hyperbolic tangent function $\tanh:\A \to \A$ is defined as follows for any $x = \hyperN{x} \in \A$:
\bb \label{eq:split-tanh} 
\tanh(x) = \tanh(x_0) + \tanh(x_1)\ii_1\ldots + \tanh(x_n) \ii_n. \ee 
The output layer parameters are arranged in a matrix $\matM \in \A^{L \times M}$. In ELM models, this layer usually consists of a linear combination with no activation function. Hence, the output of the ELM defined on the hypercomplex algebra is obtained simply by means of the vector-matrix product 
\bb \label{eq:output-layer} \vety = \boldsymbol{h} \matM \in \A^{M}. \ee

As previously stated, real-valued ELMs use randomly generated hidden layer parameters while the output layer parameters are adjusted using a least-squares problem. Analogously, the hypercomplex-valued ELM is initiated by randomizing the hidden layer parameters. Training the output layer parameters is done by solving a hypercomplex-valued least squares problem (see Definition \ref{def:hypercomplex-LSP}).

Formally, consider a training set of hypercomplex-valued input-target pairs of the form $\mathcal{T} = \{(\vetx_i,\vett_i):i=1,\ldots,M\} \subset \A^D \times \A^N$. We organize the $M$ training elements as rows in matrices $\matX \in \A^{M \times D}$ and $\matT \in \A^{M \times N}$. In other words, the $i$-th row in matrix $\matX$ contains the input associated to the target in the $i$-th row of matrix $\matT$. The hidden layer contains $L$ neurons and is therefore represented by a randomly generated matrix $\matW \in \A^{D\times L}$. We consider a random parameter of the form
\bb w_{ij} = \alpha( \mathtt{randn} + \mathtt{randn}\ii_1 + \ldots + \mathtt{randn} \ii_n),\ee 
where $\alpha$ is a scaling factor and $\mathtt{randn}$ yields a random number drawn from a normal distribution with mean $0$ and variance $1$. At this point, we would like to recall that the activation functions are generally non-linear monotonic limited functions, therefore horizontally asymptotic. As a consequence, sufficiently large numbers in the domain are mapped into very similar values in the image, essentially degenerating the ability of the model to distinguish inputs. The purpose of the scaling factor is to avoid this effect by concentrating the values around the sensitive area of $f$. In our implementations, knowing \textit{a priori} the entries $x_i = \hyperN{x_i}$, $i=1,\ldots,D$, of an input $\vetx$ satisfy $-1 \leq x_{ij} \leq +1$ for all $i$ and $j=1,\ldots,n$, we used $\alpha = 10/D$. 

Finally, we obtain the parameters of the output layer by solving the hypercomplex-valued least squares problem
\bb \label{eq:CayleyLSP-ELM} \min\{\| \matH \matM - \matT \|_F: \matM \in \A^{L \times M}\},\ee
where $\matH = f(\matX \matW)$ is the hidden layer output matrix of the neural network. From \eqref{eq:solution-hypercomplexLS}, we have 
\bb \label{eq:CayleyLSP-solution} \matM = \varphi^{-1}\left(\Phi_L(\matH)^\dag \varphi(\matT) \right),\ee
where $\Phi_L$ and $\varphi$ are the operators defined respectively by \eqref{eq:Matrix-Phi_L} and \eqref{eq:Matrix-varphi} and $\Phi_L(\matH)^{\dagger}$ is the pseudoinverse of $\Phi_L(\matH)$. Once again we note that \eqref{eq:CayleyLSP-solution} depends on the multiplication tables of the algebra $\A$ due to the usage of $\Phi_L$.

\section{Two Applications on Four-Dimensional Algebras}
%\section{Computational Experiments - Color Image Auto-Encoder}
\label{sec:Applications}

One of the key aspects of this work is comparing the proposed ELM model based on different hypercomplex algebras of the same dimension. As previously stated, hypercomplex neural network models are well adapted to tasks involving high-dimensional data, that is, when the input consists of many signals related to the same object. For that purpose, we carried out two experiments: one featuring a time-series prediction task and one involving color image auto-encoding. Both experiments include four-dimensional hypercomplex-valued ELMs based on the seven algebras described in the following subsection. We also consider a real-valued ELM of equivalent size for comparison purposes.

The total number of parameters (TNP), defined as the sum of the total number of free parameters of a network, has been used to work with comparable size networks. In the case of ELMs, this includes the number of randomly initialized fixed parameters. All ELMs were taken as single hidden layer networks with bias terms added in both hidden layer and output layers. Formally, a real-valued network with an input signal of dimension $D$, $L$ neurons in the hidden layer, and output of dimension $O$, referred to as $D-L-O$ real-valued network, has \begin{equation} \label{eq:TNP}
    TNP = (D+1)L + (L+1)O.
\end{equation}
A four-dimensional hypercomplex-valued ELM has $4$-times the TNP of the real-valued with the same layout.

\subsection{Seven Notable Four-Dimensional Hypercomplex Algebras} \label{subsec:notable_algebras}

In this subsection, we review seven four-dimensional hypercomplex algebras. Precisely, we provide their multiplication table and address their main properties. To simplify the notation, we denote the hyperimaginary units by $\ii \equiv \ii_1$, $\jj \equiv \ii_2$, and $\kk \equiv \ii_3$. Thus, an element of a four-dimensional hypercomplex algebra is generally represented by 
\bb \label{eq:hyper4} 
x = x_0 + x_1 \ii + x_2 \jj + x_3 \kk.
\ee
The main difference between the seven hypercomplex algebras resides in their corresponding multiplication table. 

\begin{example}[Quaternions]

Quaternions ($\Q$) are arguably one of the most well-known hypercomplex algebras. Introduced in 1843 by W. R. Hamilton, quaternions are an extension of complex numbers and represent three-dimensional space rotations compactly. The product of quaternion hyperimaginary units is anticommutative and satisfies Table \ref{tab:four-dimension-alg}.

% \begin{table}
%     \caption{Multiplication table of Quaternions.}
%     \label{tab:quat}
%     \begin{center}
%         \begin{tabular}{c|rrr} 
%              $\Q$ & $\ii$ & $\jj$ & $\kk$ \\ \hline
%             $\ii$ & $-1$ & $\kk$ & $-\jj$ \\  
%             $\jj$ & $-\kk$ & $-1$ & $\ii$ \\  
%             $\kk$ & $\jj$ & $-\ii$ & $-1$\\  
%         \end{tabular}    
%     \end{center}
% \end{table}

\end{example}

\begin{example}[Cayley-Dickson Algebras]
% The octonions are eight-dimensional hypercomplex numbers. Although they are also called Cayley numbers, octonion algebra has been proposed independently by A. Cayley and J. Graves. Inspired by the works of Cayley,
Dickson developed in 1919 a recursive process that generates algebras of doubling dimension \cite{shafer54}. Using this recursive process, complex numbers are obtained from real numbers. Similarly, quaternions are obtained from complex numbers.
Octonions, also known as Cayley numbers, are obtained from quaternions \cite{culbert07}. Algebras generated by Dickson's recursive process are called Cayley-Dickson algebras. In this paper, we consider a generalized version of the Cayley-Dickson algebras proposed by Albert in 1942 \cite{albert42}. 
%Appendix \ref{appendix:Cayley-Dickson} details the recursive process used to define the Cayley-Dickson algebra. Furthermore, 
A detailed account of ELMs defined on Cayley-Dickson algebras over $\R$ can be found in \cite{Vieira2020ExtremeAuto-Encoding}.  

For the applications present in this section, we consider the four-dimensional Cayley-Dickson algebras over $\R$. As pointed out previously, quaternions are an example of a four-dimensional Cayley-Dickson algebra over $\R$, denoted by $\R[-1,-1]$. The other three notable four-dimensional Cayley-Dickson algebras over $\R$ are $\R[+1,+1]$ (hyperbolic quaternions), $\R[-1,+1]$ (coquaternions or split-quaternions), and $\R[+1,-1]$. Their multiplication tables are shown in Table \ref{tab:four-dimension-alg}.

% In this work, we shall focus on the four-dimensional algebras obtained through this process, specifically the ones described in \cite{Vieira2020ExtremeAuto-Encoding}. Quaternions, in the previous example, are one such algebra. Other three notable four-dimensional algebras over $\R$ are $\R[+1,+1]$ (hyperbolic quaternions), $\R[-1,+1]$ (coquaternions or split-quaternions) and $\R[+1,-1]$. Their multiplication tables are shown in Table \ref{tab:four-dimension-alg}.

\begin{table*}
\centering
    \caption{Multiplication tables of four-dimensional Cayley-Dickson algebras, including the quaternions $(\Q \thickapprox \R[-1,-1])$.}
    \label{tab:four-dimension-alg}
\begin{tabular}{rl}
\begin{tabular}{c|rrr} 
             $\Q$ & $\ii$ & $\jj$ & $\kk$ \\ \hline
            $\ii$ & $-1$ & $\kk$ & $-\jj$ \\  
            $\jj$ & $-\kk$ & $-1$ & $\ii$ \\  
            $\kk$ & $\jj$ & $-\ii$ & $-1$\\  
        \end{tabular}    & $\quad$ 
\begin{tabular}{c|rrr} 
         $\R \left[+1,+1\right]$ &  $\ii$ & $\jj$ & $\kk$ \\ \hline
         $\ii$ & $1$ & $\kk$ & $\jj$ \\  
         $\jj$ & $-\kk$ & $1$ & $-\ii$ \\ 
         $\kk$ & $-\jj$ & $\ii$ & $-1$\\  
    \end{tabular}    
    \\  & \\ 
\begin{tabular}{c|rrr}
     $\R\left[-1,+1 \right]$ & $\ii$ & $\jj$ & $\kk$ \\ \hline
     $\ii$ & $-1$ & $\kk$ & $-\jj$ \\  
     $\jj$ & $-\kk$ & $1$ & $-\ii$ \\  
     $\kk$ & $\jj$ & $\ii$ & $1$\\  
    \end{tabular}
    & $\quad$ 
\begin{tabular}{c|rrr}
         $\R\left[ +1,-1 \right]$ & $\ii$ & $\jj$ & $\kk$ \\ \hline
         $\ii$ & $1$ & $\kk$ & $\jj$ \\  
         $\jj$ & $-\kk$ & $-1$ & $\ii$ \\  
         $\kk$ & $-\jj$ & $-\ii$ & $1$\\  
    \end{tabular} 
\end{tabular}
\end{table*}  
\end{example}

The Cayley-Dickson process is known for losing properties with each increment in dimension. In contrast, the Clifford algebras are equipped with operations with elegant geometric interpretations \cite{hestenes12}.

\begin{example}[Clifford Algebras]
Briefly, a Clifford algebra is an associative algebra generated by a vector space equipped with a quadratic form. Their operations' geometric properties play an important role in many applications, which range from physics \cite{crumeyrolle13,chisholm96} to digital signal processing \cite{labunets04}.

A Clifford algebra over $\R$ is denoted by $C\ell _{p,q}(\R)$, or simply $C\ell_{p,q}$, where $p$ and $q$ are non-negative integers.
% This notation specifies that the generating vector space has an orthogonal basis with $p$ elements such that $\e{i}^2 = +1$ and $q$ elements such that $\e{j}^2 = -1$. One important property of one such algebra is that
% $$ \e{j} \e{i} = - \e{i} \e{j}, \quad i \neq j. $$
% Letting $n=p+q$, a basis for $C\ell _{p,q}$ has the form
% $$ \left\lbrace \e{i_1}\e{i_2}\dots \e{i_k} \quad | \quad 1<i_1<\dots<i_k<n, \quad 0\leq k \leq n \right\rbrace, $$
% i.e., consists of all combinations of up to $n$ elements in ascending index order. It follows that
% \bb 
% \dim C\ell _{p,q}(\R) = 2^{p+q}
% \ee
The dimension of the Clifford algebra $C\ell_{p,q}$ is $n = 2^{p+q}$. Because we are interested on four-dimensional Clifford algebras, we only consider $C\ell_{p,q}$ such that $p+q=2$.
The algebra $C\ell_{0,2}$ is equivalent to the quaternions, whose multiplication table is given by Table \ref{tab:four-dimension-alg}. 
% For $(p,q) = (0,2)$ we are left with a four-dimensional algebra with units $\lbrace 1, \e{2}, \e{3}, \e{2}\e{3} \rbrace$. The three non-identity elements have squares equal to $-1$ and the product anticommutes, thus by relabeling $\e{2} = \ii$, $\e{3} = \jj$ and $\e{2}\e{3} = \kk$ it is clear that $C\ell_{0,2} \thickapprox \Q $, i.e., this algebra is equivalent to the quaternions and their muliplication tables are the same (Table \ref{tab:quat}). 
The two remaining configurations are $C\ell_{1,1}$ and $C\ell_{2,0}$, both of which are equivalent to $M_2(\R)$, the algebra of square real matrices of order $2$ \cite{porteous95}. Table \ref{tab:non_cd_algebras} contains the multiplication table for $C\ell_{1,1} \thickapprox C\ell_{2,0}$.
%\begin{table}
%    \caption{Multiplication table of $C\ell_{1,1}$ and $C\ell_{2,0}$.}
%    \label{tab:m2}
%    \begin{center}
%        \begin{tabular}{c|rrrr} 
%             $C\ell_{1,1/2,0}$ & $1$ & $\e{2}$ & $\e{3}$ & $\e{4}$ \\ %\hline
%            $1$ & $1$ & $\e{2}$ & $\e{3}$ & $\e{4}$ \\
%            $\e{2}$ & $\e{2}$ & $1$ & $-\e{4}$ & $-\e{3}$ \\  
%            $\e{3}$ & $\e{3}$ & $\e{4}$ & $1$ & $-\e{2}$ \\  
%            $\e{4}$ & $\e{4}$ & $\e{3}$ & $\e{2}$ & $1$\\  
%        \end{tabular}    
%    \end{center}
%\end{table}
\end{example}

Commutativity is a key property for algebras, especially from a computational standpoint: it ensures many simplifications and reduces operations' computational complexity. Despite being associative, the Clifford algebras, including the quaternions, usually fail to be commutative. Similarly, Cayley-Dickson algebras often lack the commutative property. The following two examples address commutative four-dimensional algebras.

\begin{example}[Tessarines]
The tessarines, introduced in 1849 and also known as bicomplex numbers \cite{rochon04}, is a commutative four-dimension algebra that differs slightly from the Cayley-Dickson algebra $\R [-1,+1]$ (split-quaternions). Table \ref{tab:non_cd_algebras} shows the multiplication for tessarines. Like the quaternions, tessarines have been used for digital signal processing \cite{pei04,alfsmann06}.

%\begin{table}[h]
%    \caption{Multiplication table of Tessarines.}
%    \label{tab:tessarines}
%    \begin{center}
%        \begin{tabular}{c|rrrr} 
%             $\mathbb{T}$ & $1$ & $\ii$ & $\jj$ & $\kk$ \\ \hline
%            $1$ & $1$ & $\ii$ & $\jj$ & $\kk$ \\
%            $\ii$ & $\ii$ & $-1$ & $\kk$ & $-\jj$ \\  
%            $\jj$ & $\jj$ & $\kk$ & $1$ & $\ii$ \\  
%            $\kk$ & $\kk$ & $-\jj$ & $\ii$ & $-1$\\  
%        \end{tabular}    
%    \end{center}
%\end{table}
\end{example}

\begin{example}[Klein Four-Group]
The Klein four-group $\K$ is a four-dimension algebra whose hyperimaginary unit are self-inverse, i.e., $\ii^2 = \jj^2 = \kk^2 = 1$. Furthermore, the product of two hypercomplex units results in the third. The multiplication table of the Klein four-group is depicted in Table \ref{tab:non_cd_algebras}. Besides the theoretical studies in symmetric group theory \cite{huang13klein,craven11}, the Klein four-group has been used for the design of hypercomplex-valued Hopfield neural networks \cite{Kobayashi2020HopfieldFour-group}.

%\begin{table}[h]
%    \caption{Multiplication table of the Klein four-group.}
%    \label{tab:klein}
%    \begin{center}
%        \begin{tabular}{c|rrrr} 
%             $\K$ & $1$ & $\e{2}$ & $\e{3}$ & $\e{4}$ \\ \hline
%            $1$ & $1$ & $\e{2}$ & $\e{3}$ & $\e{4}$ \\
%            $\e{2}$ & $\e{2}$ & $1$ & $\e{4}$ & $\e{3}$ \\  
%            $\e{3}$ & $\e{3}$ & $\e{4}$ & $1$ & $\e{2}$ \\  
%            $\e{4}$ & $\e{4}$ & $\e{3}$ & $\e{2}$ & $1$\\  
%        \end{tabular}    
%    \end{center}
%\end{table}
\end{example}

\begin{table*}
\centering
    \caption{Multiplication tables of the Clifford algebra ($C\ell_{1,1} \thickapprox C\ell_{2,0}$), tessariens ($\mathbb{T})$, and Klein four-group ($\K$).}
    \label{tab:non_cd_algebras}
\begin{tabular}{lcr}
\begin{tabular}{c|rrrr} 
             $C\ell_{1,1}$  & $\ii$ & $\jj$ & $\kk$ \\ \hline
            $\ii$ & $1$ & $-\kk$ & $-\jj$ \\  
            $\jj$ & $\kk$ & $1$ & $-\ii$ \\  
            $\kk$ & $\jj$ & $\ii$ & $1$\\  
        \end{tabular}    
    & $\quad$
    \begin{tabular}{c|rrrr} 
             $\mathbb{T}$ & $\ii$ & $\jj$ & $\kk$ \\ \hline
            $\ii$ & $-1$ & $\kk$ & $-\jj$ \\  
            $\jj$ & $\kk$ & $1$ & $\ii$ \\  
            $\kk$ & $-\jj$ & $\ii$ & $-1$\\  
        \end{tabular} & $\quad$
    \begin{tabular}{c|rrrr} 
             $\K$ & $\ii$ & $\jj$ & $\kk$ \\ \hline
            $\ii$ & $1$ & $\kk$ & $\jj$ \\  
            $\jj$ & $\kk$ & $1$ & $\ii$ \\  
            $\kk$ & $\jj$ & $\ii$ & $1$\\  
        \end{tabular}
\end{tabular} 
\end{table*}  

% \begin{remark}
% Note in Table \ref{tab:non_cd_algebras} that both Klein four-group and tessarines have symmetric multiplication tables because they are commutative algebras.
% \end{remark}

% We are therefore provided with \textbf{seven} four-dimensional hypercomplex algebras with different properties:
% \begin{itemize}
%     \item Four Cayley-Dickson algebras, including the quaterions;
%     \item The Clifford algebra which are associative, anticommutative and share the same multiplication table;
%     \item the Tessarines, a commutative algebra similar to one of the Cayley-Dickson configurations;
%     \item the Klein four-group, a commutative algebra with all imaginary units' products having a positive sign.
% \end{itemize}

% \begin{remark} \label{remark:equivalence}
% As seen from \eqref{eq:hc-number}, an element in a hypercomplex algebra over $\R$ can be interpreted as a linear combination of the hypercomplex basis units by real parameters. As so, it can be identified uniquely with the $(n+1)$-tuple
% \bb p_0 + p_1 \e{1} + \dots + p_n \e{n} \longleftrightarrow (p_0,p_1,\dots,p_n) \in \R^{n+1}. \ee
% With addition and scalar product defined in \eqref{eq:hc-addition}-\eqref{eq:scalar_product}, this equivalence means that hypercomplex algebras can be identified with Euclidean vector space structures.
% \end{remark}

We evaluate the real-valued and hypercomplex-valued ELM models' performance using the seven hypercomplex algebras for application tasks in the following subsection. 

\subsection{Times Series Prediction}

For the time series prediction task, we considered the well known Lorenz system. The Lorenz system is a chaotic system of ordinary differential equations describing a nonperiodic flow on a three dimensional space. Formally, the system is given by
\begin{equation} \label{eq:lorenz_sys}
\begin{cases}
    \frac{dx}{dt}  = \sigma (y-x),  \\
    \frac{dy}{dt}  = x(\rho - z) - y, \\
    \frac{dz}{dt}  = xy - \beta z, 
\end{cases}
\end{equation}
where the variables $(x,y,z)$ corresponds to a position in space, and $\sigma, \rho, \beta>0$ are constants. The Lorenz system is chaotic for some constant values, meaning that small perturbations in the initial condition often incur large changes in the result. We considered $\sigma = 10$, $\beta = 8/3$, and $\rho = 28$ in our computational experiments.

As usual in time series prediction \cite{Datar2002MaintainingWindows,xu16}, we consider a sliding window of fixed length $T$. In other words, $T$ consecutive positions are used as input for a model that attempts to predict the next position. In our experiment, we used $T=3$. Thus, an arbitrary training sample has the positions $\mathbf{p}_{t-2}=(x_{t-2},y_{t-2},z_{t-2})$, $\mathbf{p}_{t-1}=(x_{t-1},y_{t-1},z_{t-1})$, and $\mathbf{p}_{t}=(x_t,y_t,z_t))$ as input while the desired output is the position $\mathbf{p}_{t+1}=(x_{t+1},y_{t+1},z_{t+1})$. The real-valued network has $9$ real input values, obtained by concatenating the all the variables. In mathematical terms, the input of the real-valued ELM is of the form:
\begin{itemize}
    \item \textbf{Real input:} $(x_{t-2},y_{t-2},z_{t-2},x_{t-1},y_{t-1},z_{t-1},x_t,y_t,z_t)$.
\end{itemize}
The desired output for the real-valued model is simply the 3-position vector $\mathbf{p}_{t+1}$.
Each position is encoded in the hyperimaginary part of distinct numbers in the four-dimensional hypercomplex-valued models. As a consequence, the input of an hypercomplex-valued ELM is of the form
\begin{itemize}
    \item \textbf{Hypercomplex input:} $(p_{t-2},p_{t-1},p_{t})$,
\end{itemize}
where $p_{t} = x_t \ii + y_t \jj + z_t \kk$ for discrete time instant $t$.
% Summarizing, the inputs have the following forms:
% \begin{itemize}
%     \item \textit{Real-valued model:} \[(x_{k-2},y_{k-2},z_{k-2},x_{k-1},y_{k-1},z_{k-1},x_k,y_k,z_k).\]
%     \item \textit{Hypercomplex-valued models:} 
%     \[ (x_{k-2} \ii + y_{k-2} \jj + z_{k-2} \kk, x_{k-1} \ii + y_{k-1} \jj + z_{k-1} \kk, x_k \ii + y_k \jj + z_k \kk). \]
%     % \begin{equation} \begin{split} (&(0 + x_0 \ii_1 + y_0 \ii_2 + z_0 \ii_3), \\ & (0 + x_1 \ii_1 + y_1 \ii_2 + z_1 \ii_3), \\ & (0 + x_2 \ii_1 + y_2 \ii_2 + z_2 \ii_3)) \end{split} \end{equation}
% \end{itemize}
The output of the hypercomplex-valued network is the hypercomplex number $p_{t+1} = x_{t+1}\ii + y_{t+1}\jj + z_{t+1}\kk$.

A total of $4.000$ consecutive positions were generated using a fourth-order Runge-Kutta method. The first $300$ positions have been used for training, while the remaining $3.700$ positions have been used for testing. Because of the sliding window, the training and test sets have 297 and 3697 samples, respectively.

We used the prediction gain to evaluate the performance of the ELM models. The prediction gain is defined by
\begin{equation} \label{eq:pred_gain}
    R = 10 \log_{10} \frac{\sigma_s^2}{\sigma_e^2},
\end{equation}
where $\sigma_s^2$ is the estimated variance of the input signal and $\sigma_e^2$ denotes the estimated variance of the prediction error. Precisely, we computed the sample variances
\begin{equation}
    \sigma_s^2 = \frac{1}{N_S-1} \sum_{t=1}^{N_S} (\|\mathbf{p}_t\|_2 - \mu_s)^2, 
\end{equation}
and
\begin{equation}
    \sigma_e^2 = \frac{1}{N_S-1} \sum_{t=1}^{N_S} (\|\mathbf{p}_{t+1} - \hat{\mathbf{p}}_{t+1}\|_2 - \mu_e)^2,
\end{equation}
where $\hat{\mathbf{p}}_{t+1} = \mathtt{ELM}(\mathbf{p}_{t-2},\mathbf{p}_{t-1},\mathbf{p}_{t})$ is the position predicted by an ELM model, $N_S$ is the number of samples, and 
\begin{equation} 
\mu_s = \frac{1}{N_S} \sum_{t=1}^{N_S} \|\mathbf{p}_t\|_2 
\quad \text{and} \quad 
\mu_e = \frac{1}{N_S} \sum_{t=1}^{N_S} \|\mathbf{p}_{t+1} - \hat{\mathbf{p}}_{t+1}\|_2. 
\end{equation}
are the input and error means, respectively.
%This definition encompasses cases in which the input and output signals are one-dimensional. In our case, we considered the 2-norm to represent the 3-dimensional input signal and error. Thus, the $\sigma$'s in \eqref{eq:pred_gain} are taken over a vector of 2-norms of the respective quantities.

%For the network hidden layer sizes we took the approach of leveling by the total number of parameters (TNP). In each of the models a bias factor was added in the input and hidden layer, the bias for the hypercomplex models is $1+0\ii_1+0\ii_2+0\ii_3$. A real-valued network with $L^{({\R})}$ neurons in the hidden layer, $9-L^{({\R})}-3$, contains $TNP^{({\R})} = 13 L^{({\R})} + 3$. A hypercomplex model with $L^{(\mathbb{H})}$ neurons in the hidden layer, $3-L^{(\mathbb{H})}-1$, contains $TNP^{(\mathbb{H})} = 20 L^{(\mathbb{H})} + 4$. 

According to \eqref{eq:TNP}, the TNP for the real-valued ELM with $L^{(\R)}$ hidden neurons is $\text{TNP}^{({\R})} = 13 L^{({\R})} + 3$. For a hypercomplex-valued network with $L^{(\mathbb{H})}$ hidden neurons, $\text{TNP}^{(\mathbb{H})} = 20 L^{(\mathbb{H})} + 4$. Imposing $\text{TNP}^{(\R)} \simeq \text{TNP}^{(\mathbb{H})}$, we obtain the following relationship for the number of hidden neurons:
\begin{equation} \label{eq:LH2LR}
    L^{(\R)} \simeq \frac{20 L^{(\mathbb{H})}}{13}.
\end{equation}

\begin{figure}[t]
    \centering
    \includegraphics[width=\textwidth]{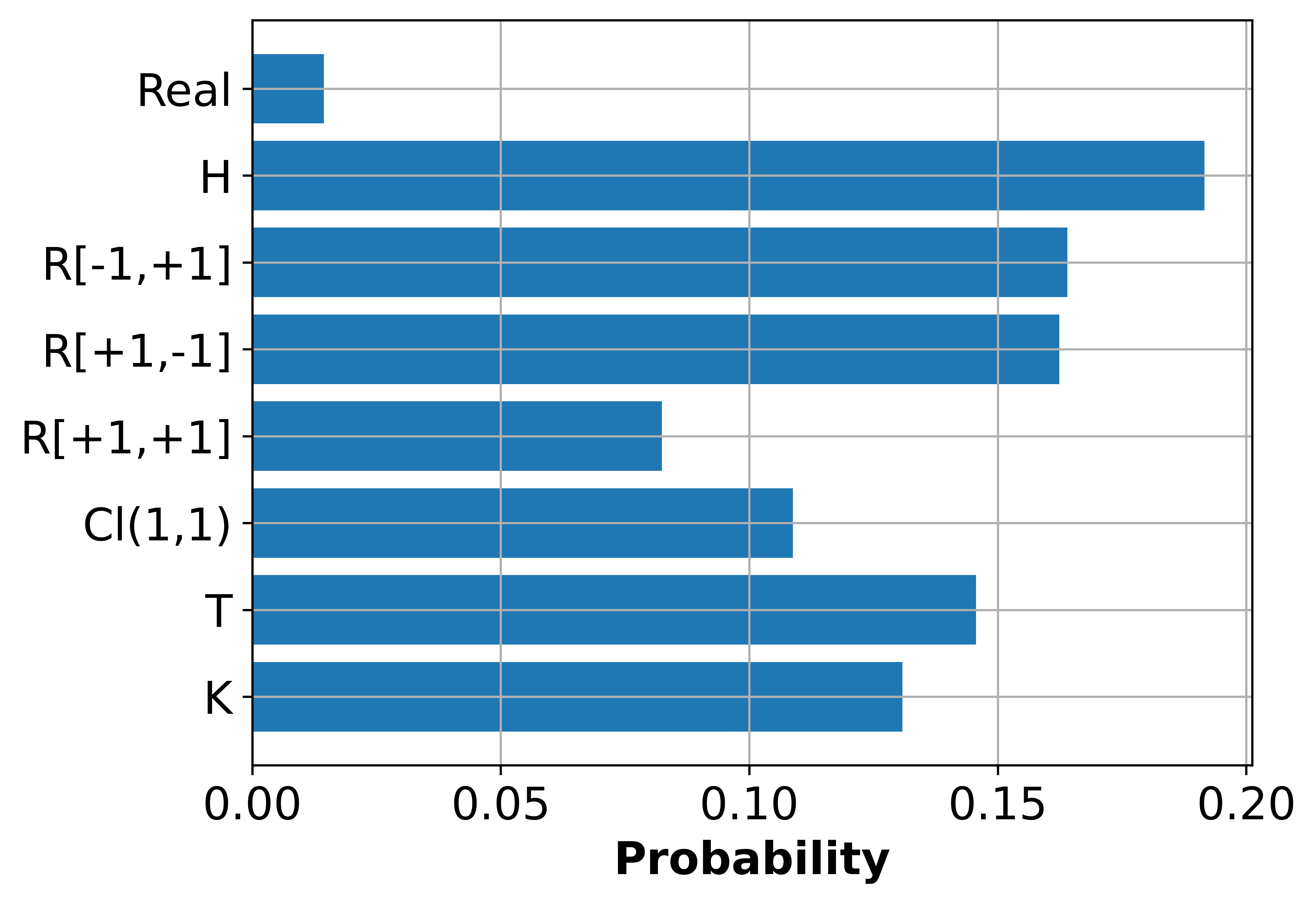}
    \caption{The probability of an ELM predictor yields the highest prediction gain by the underlying algebra.}
    \label{fig:lorenz_sims}
\end{figure}

We performed a series of tests with $L^{(\mathbb{H})}$ ranging in $\lbrace 11, 12, \dots, 34, 35 \rbrace$ and determined the corresponding number of hidden neurons for the real-valued ELM using \eqref{eq:LH2LR}. For each value of $L^{(\mathbb{H})}$, a total of 100 networks have been trained for each algebra, resulting in a total of $20.000$ simulations. For each of these simulations, we annotated the best performing model, i.e., the model that showcased the highest prediction gain. Fig. \ref{fig:lorenz_sims} shows the probability of an ELM model yield the highest prediction gain in the 20.000 simulations.
%The results were then normalized by the total number of simulations (100). The results are shown in Figure \ref{fig:lorenz_sims}.

\begin{figure}[t]
    \includegraphics[width=\textwidth]{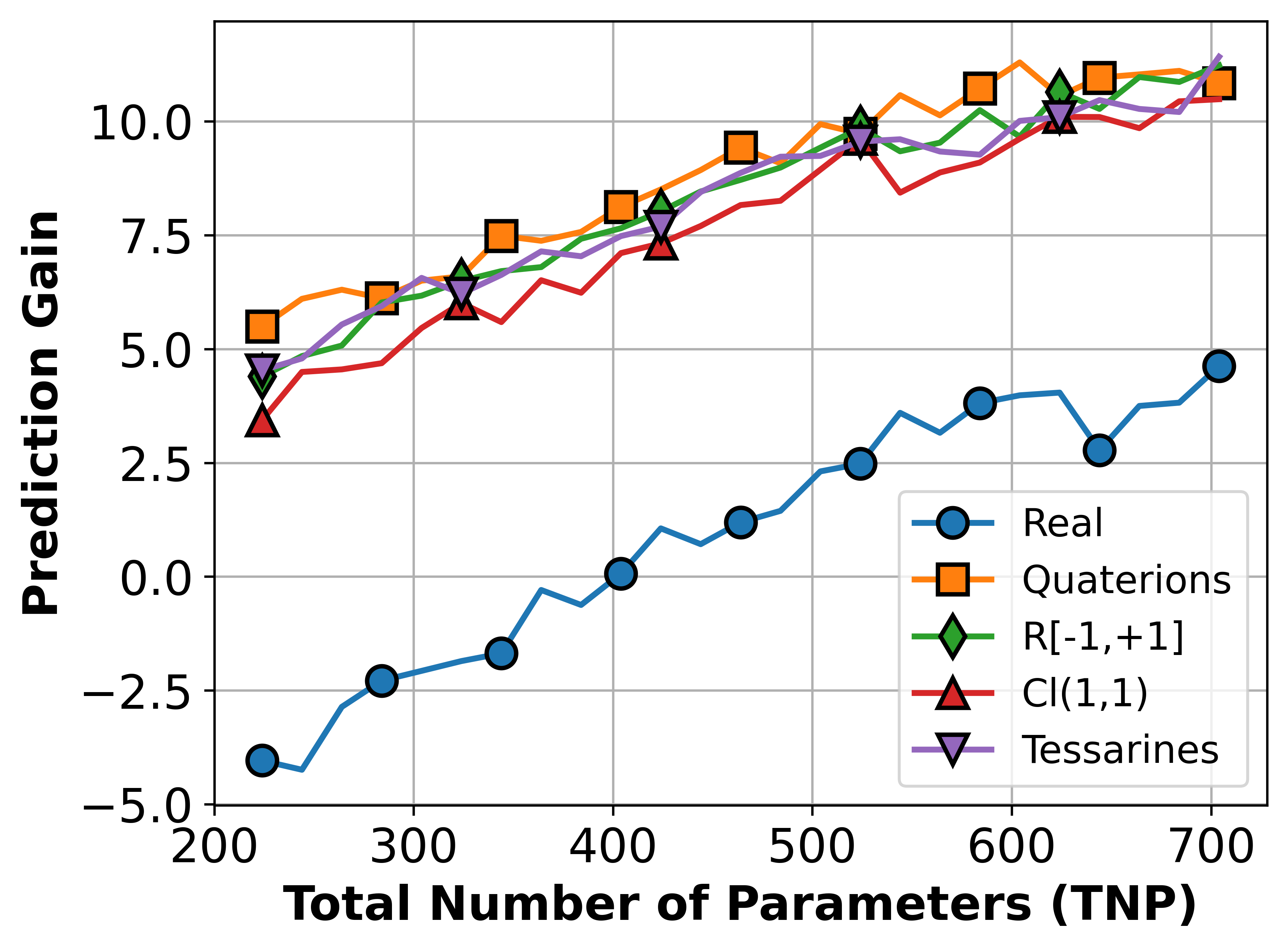}
    \caption{The average prediction gain by the total number of parameters.}
    \label{fig:predgains}
\end{figure}

At a glance, it is clear that the real-valued model underperformed when compared to the hypercomplex-valued ones, seldom showcasing the highest prediction gain. Moreover, the three top-performing models were Cayley-Dickson algebras, namely the quaternions (H, $\R[-1,-1])$, followed by $\R[-1,+1]$ and $\R[+1,-1]$. The hypercomplex models' advantage over the real-valued becomes clearer by taking the average prediction gain over 100 simulations for each number of hidden neurons. Fig. \ref{fig:predgains} shows the average prediction gain by the total number of parameters for the real-valued ELM and three representative hypercomplex-valued models. Namely, the ELM based on the quaternions, the Cayle-Dickson algebra $\R[-1,+1]$, the Clifford algebra $C\ell_{1,1}$, which is associative, and the tessarines, which is commutative.

\subsection{Color Image Auto-Encoding}

For the color image auto-encoding task, we considered the CIFAR-10 dataset. The CIFAR-10 dataset was originally conceived as a natural image dataset for classification tasks. It contains five training batches and one testing batch. Each batch contains $10,000$ images, amounting to a total of $60,000$ images divided evenly into $10$ classes. %Each image is encoded in a $32 \times 32$ $8$-bit RGB matrix.

Although the CIFAR-10 has been originally conceived for classification tasks, it has also been used for image auto-encoding \cite{tang16,zhang17p,minemoto17}. The task of auto-encoding consists of obtaining a model capable of compressing a high-dimensional object -- in this case, an image -- and further reconstructing the original object from the compressed information with minimal loss. This is a vital task in information theory, where the exchange of compressed minimal-loss information is invaluable. A neural network designed to perform an auto-encoding task, known as an auto-encoder, is trained using a set in which the input and the output are the same images. Moreover, the dimension of the intermediate layers is lesser than that of the image. This forces the network to learn a representation for a high-dimensional object in the space of a smaller dimension while maintaining the amount of information. For their ability to learn representations, auto-encoders are known as powerful feature detectors and are used in unsupervised pre-training on large datasets. In essence, there is a great similarity between auto-encoders and convolutional layers, which act as feature extractors in deep learning models. Lastly, auto-encoders have applications in generative models \cite{geron17}.

In this work, a total of $8$ models have been trained with a single batch of the CIFAR-10 and tested using the original test batch of the same dataset, i.e., the training and test sets consist of $10,000$ images each. The hypercomplex architectures used are based on the $7$ algebras described previously. We also include a traditional real-valued ELM. 

Each image in the CIFAR-10 dataset is encoded as $32 \times 32$ 8-bit RGB matrix, resulting a total of $3072$ values. An 8-bit RGB image $\mathbf{I}$ has been converted to a real-valued vector $\vetx^{(\mathbb{R})}$ of length $3072$ concatenating the pixel values in the red, green, and blue channels. Furthermore, the values were rescaled to fit the interval $[-1,+1]$. In the hypercomplex case, the image $\mathbf{I}$ was converted into a hypercomplex-valued vector $\vetx^{(\mathbb{H})}$ of length $1024$ with components
\[ x_i^{(\mathbf{H})} = \left(\frac{2\mathbf{I}_i^R}{255}-1\right)\ii+\left(\frac{2\mathbf{I}_i^G}{255}-1\right)\jj+\left(\frac{2\mathbf{I}_i^B}{255}-1\right)\kk,\]
where $\mathbf{I}_i^{R},\mathbf{I}_i^{G},\mathbf{I}_i^{B}$ represent the values of the $i$-th pixel at the red, green, and blue channels, respectively. 

Like the previous experiment, we considered real and hypercomplex-valued ELMs with similar TNP. Due to the symmetry of input and output, the real-valued and hypercomplex-valued ELM have respectively
\begin{equation} 
\text{TNP}^{(\R)} = 2D^{(\R)}L^{(\R)} \quad \text{and} \quad  \text{TNP}^{(\mathbb{H})} = 4(2D^{(\mathbb{H})}L^{(\mathbb{H})}),
\end{equation}
where $D^{(\R)} = 3072$ and $D^{(\mathbb{H})}=1024$ denote the dimension of the network's input. As reported previously by Minemoto et al. \cite{minemoto17} and further experimented by us \cite{Vieira2020ExtremeAuto-Encoding}, the number of hidden layer neurons were taken as $L^{(\R)}=600$ and $L^{(\mathbb{H})}=450$, amounting to an equal TNP for both models, that is, $\text{TNP}^{(\R)} =  \text{TNP}^{(\mathbb{H})} = 3,686,400$.

%All architectures are considered to have a single hidden layer with $L$ total neurons. In order to make the networks of comparable size we observed the total number of parameters (TNP). The TNP is defined as the sum of the total number of free trainable parameters and the number of randomly initialized fixed parameters. For a real-valued model with $L^{(\R)}$ neurons in the hidden layer, the TNP is
%$$ TNP^{(\R)} = 2D^{(\R)}L^{(\R)} $$
%where $D^{(\R)} = 3072$ stands for the dimension of the input vector, i.e., the image. For the hypercomplex ELMs we have
%$$ TNP^{(\mathbb{H})} = 4(2D^{(\mathbb{H})}L^{(\mathbb{H})}) $$
%where $L^{(\mathbb{H})}$ is the number of hidden layer neurons and $D^{(\mathbb{H})}=1024$ is the dimension of the hypercomplex input matrix. The constant $2$ is a consequence of the auto-encoding task, where the input and desired output objects having the same size, therefore the network is symmetric around the hidden layer. The constant $4$ in the expression for hypercomplex models appears because each four-dimensional parameter has four components and therefore is equivalent to four real parameters. As reported previously by \cite{minemoto17} and further experimented in \cite{Vieira2020ExtremeAuto-Encoding} the number of hidden layer parameters were taken as $L^{(\R)}=600$ and $L^{(\mathbb{H})}=450$, amounting to an equal TNP for both models, $TNP^{(\R)} =  TNP^{(\mathbf{H})} = 3,686,400$.

The hidden layer's activation function was the real-valued hyperbolic tangent for the real ELM and the split hyperbolic tangent for the hypercomplex ELMs. Furthermore, the hidden layer parameters were randomly generated according to a standard normal distribution with mean $0$ and variance $1$. To ensure the values would not saturate the activation function, the normally distributed parameters were re-scaled by a constant $\alpha$, which depends on the network's input length. Precisely, we used $\alpha^{(r)} = 30/3072$ and $\alpha^{(\mathbb{H})} = 10/1024$ for the real and hypercomplex-valued models, respectively.

\begin{figure}
    % \centering
    \hspace{-3mm} \begin{tabular}{ccc}
    a) Original & b) Real-valued & c) Quaternions \\
    \includegraphics[width=0.3\columnwidth]{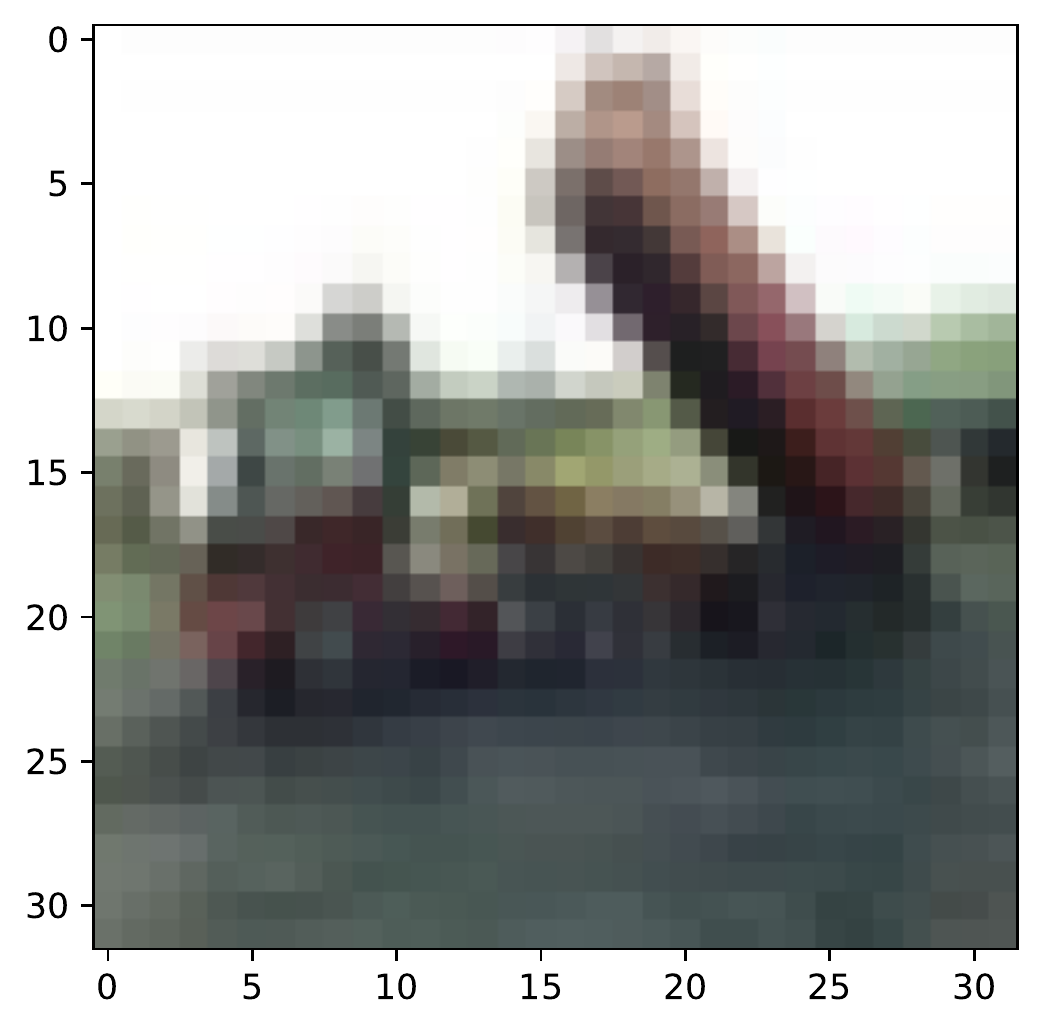} &
    \includegraphics[width=0.3\columnwidth]{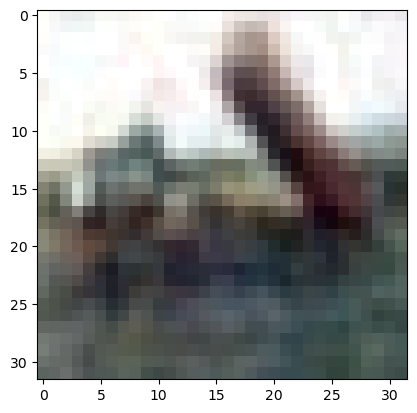} &
    \includegraphics[width=0.3\columnwidth]{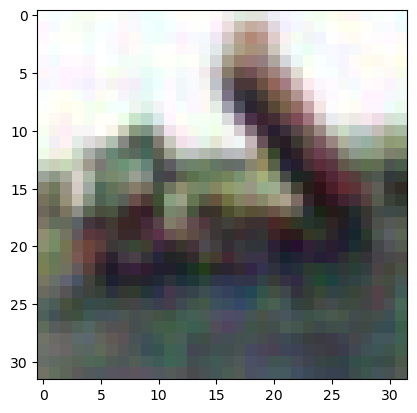} \\
    d) $\R[-1,+1]$ & e) $\R[+1,-1]$ & f) $\R[+1,+1]$ \\
    \includegraphics[width=0.3\columnwidth]{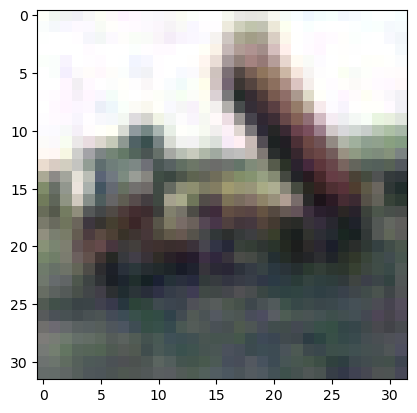} &
    \includegraphics[width=0.3\columnwidth]{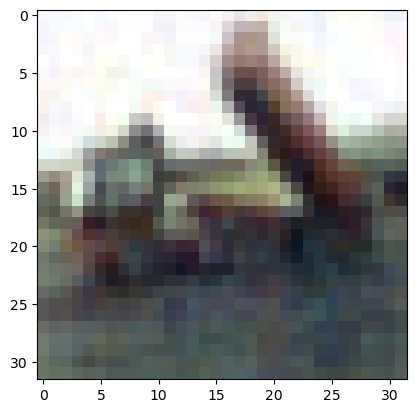} &
    \includegraphics[width=0.3\columnwidth]{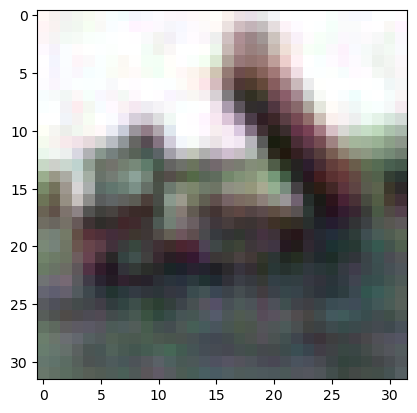} \\
    g) $C\ell_{1,1}$ & h) Klein four-group & i) Tessarines \\ 
    \includegraphics[width=0.3\columnwidth]{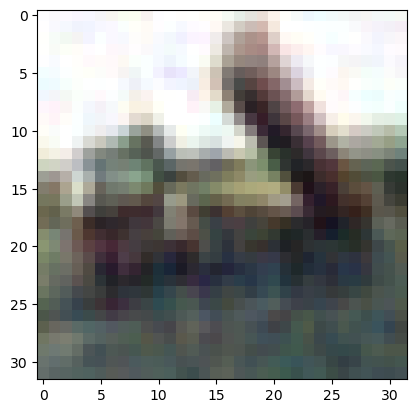} &
    \includegraphics[width=0.3\columnwidth]{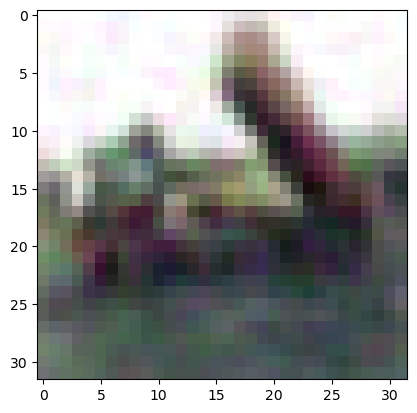} &
    \includegraphics[width=0.3\columnwidth]{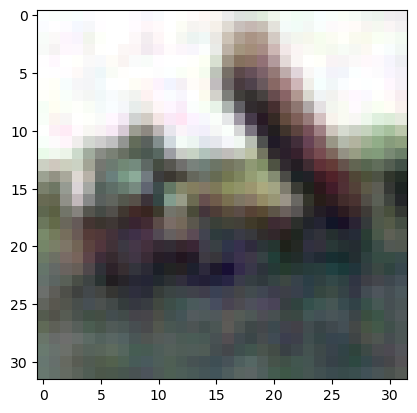}
    \end{tabular}
    \caption{Training set samples: original color images and the corresponding images decoded by the real and hypercomplex-valued auto-encoders.}
    \label{fig:images_train}
\end{figure}

\begin{figure}
    % \centering
    \hspace{-3mm} \begin{tabular}{ccc}
    a) Original & b) Real-valued & c) Quaternions \\
    \includegraphics[width=0.3\columnwidth]{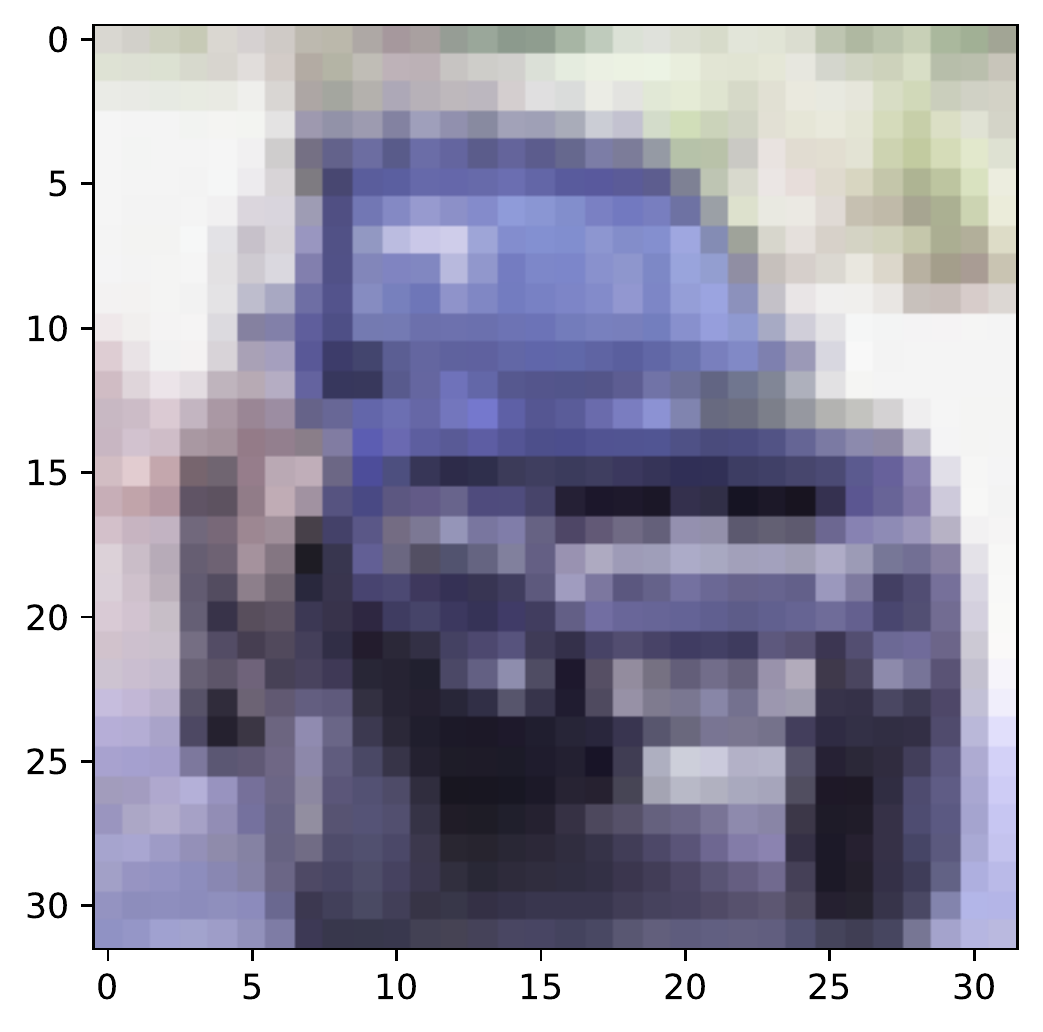} &
    \includegraphics[width=0.3\columnwidth]{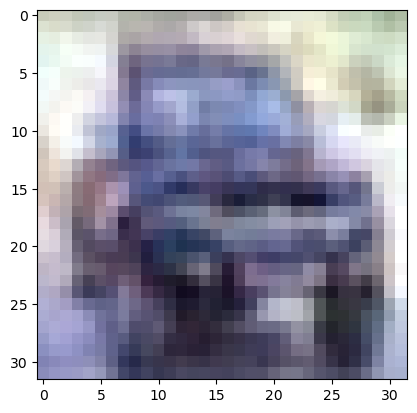} &
    \includegraphics[width=0.3\columnwidth]{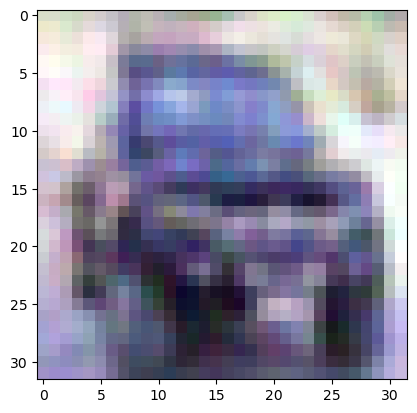} \\
    d) $\R[-1,+1]$ & e) $\R[+1,-1]$ & f) $\R[+1,+1]$ \\
    \includegraphics[width=0.3\columnwidth]{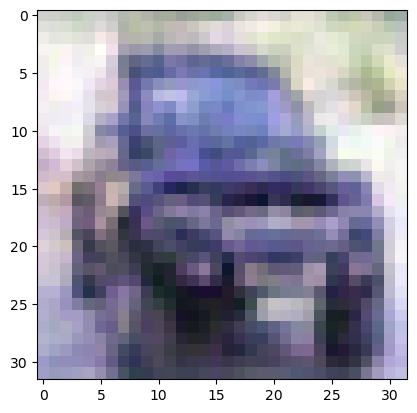} &
    \includegraphics[width=0.3\columnwidth]{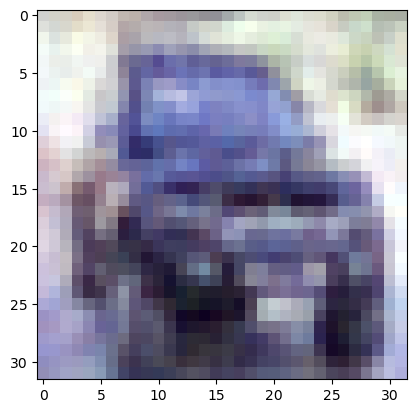} &
    \includegraphics[width=0.3\columnwidth]{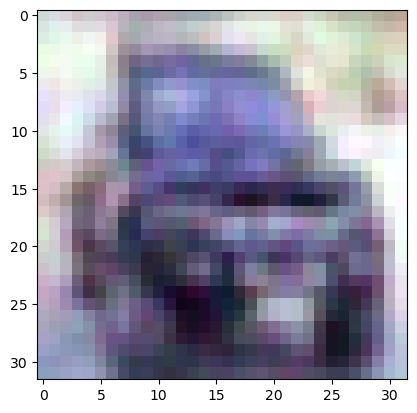} \\
    g) $C\ell_{1,1}$ & h) Klein four-group & i) Tessarines \\ 
    \includegraphics[width=0.3\columnwidth]{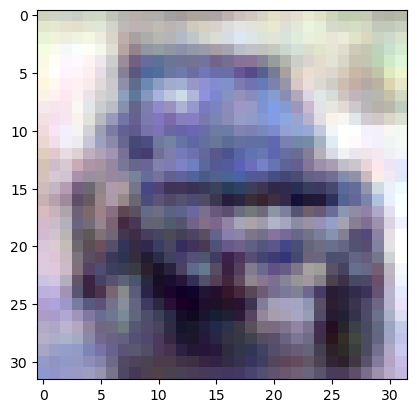} &
    \includegraphics[width=0.3\columnwidth]{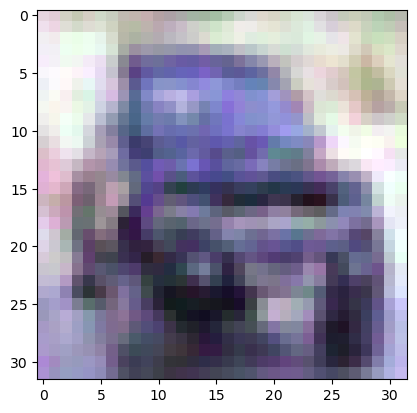} &
    \includegraphics[width=0.3\columnwidth]{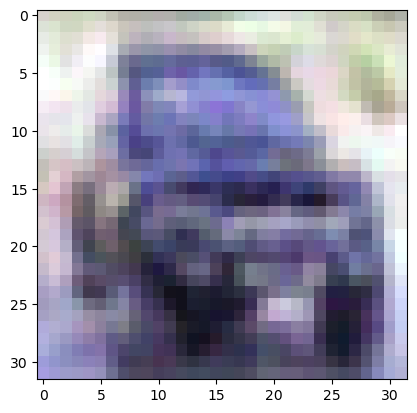}
    \end{tabular}
    \caption{Test set samples: original color images and the corresponding images decoded by the real and hypercomplex-valued auto-encoders.}
    \label{fig:images_test}
\end{figure}

The first training batch of the CIFAR-10 dataset, containing 10,000 images, has been used to train the eight ELM auto-encoders. The test batch, containing 10,000 different images, was used for testing. For illustrative purposes, Figs. \ref{fig:images_train} and \ref{fig:images_test} show the original color images from the CIFAR dataset and the corresponding images decoded by the real and hypercomplex-valued auto-encoders. Precisely, Fig. \ref{fig:images_train} shows the results for a training image while Fig. \ref{fig:images_test} depicts the outcome of a testing image.

\begin{figure}
    \centering
    % \begin{tabular}{c}
        a)  PSNR  \\
        \includegraphics[width=1\columnwidth]{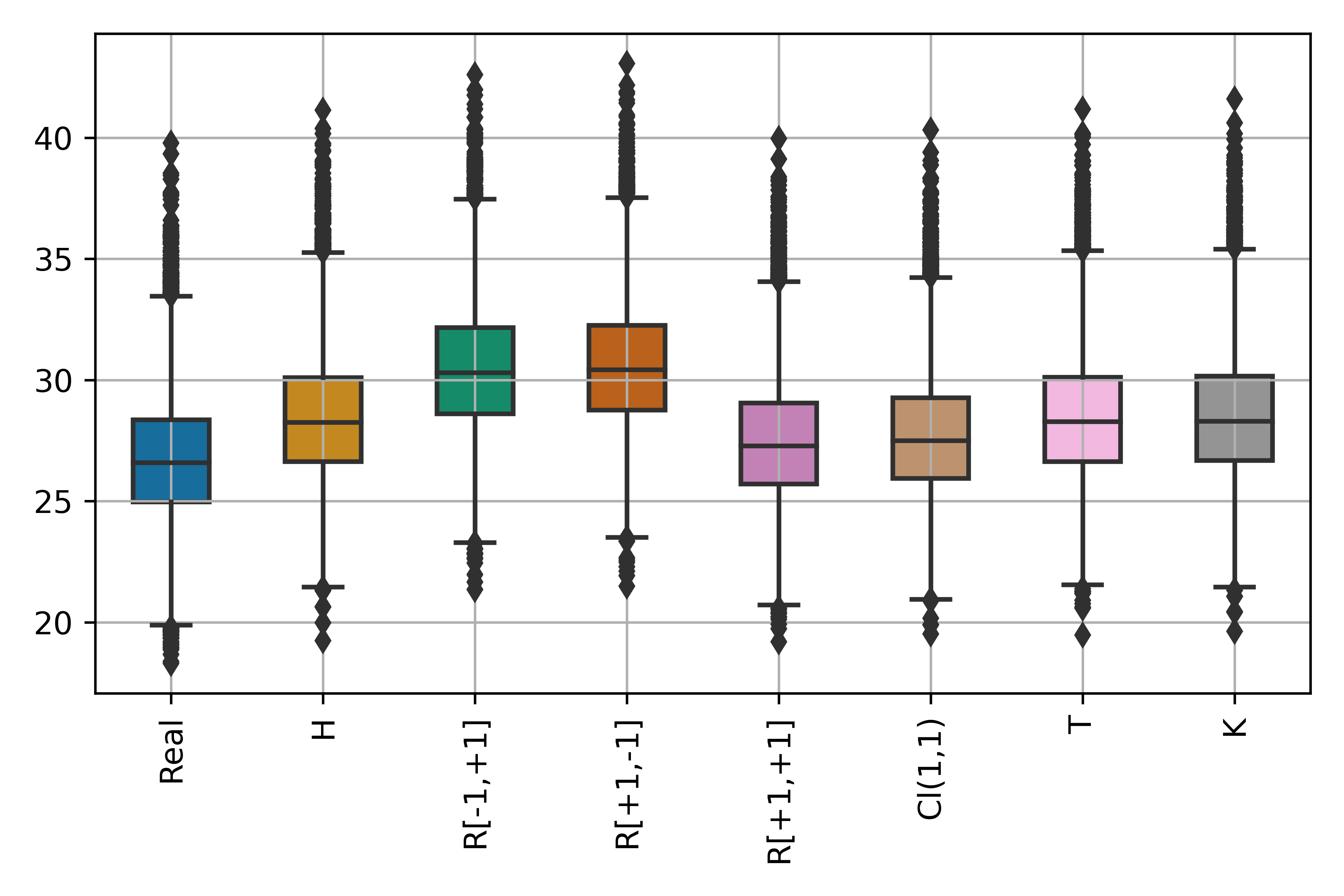} \\ b) SSIM \\
        \includegraphics[width=1\columnwidth]{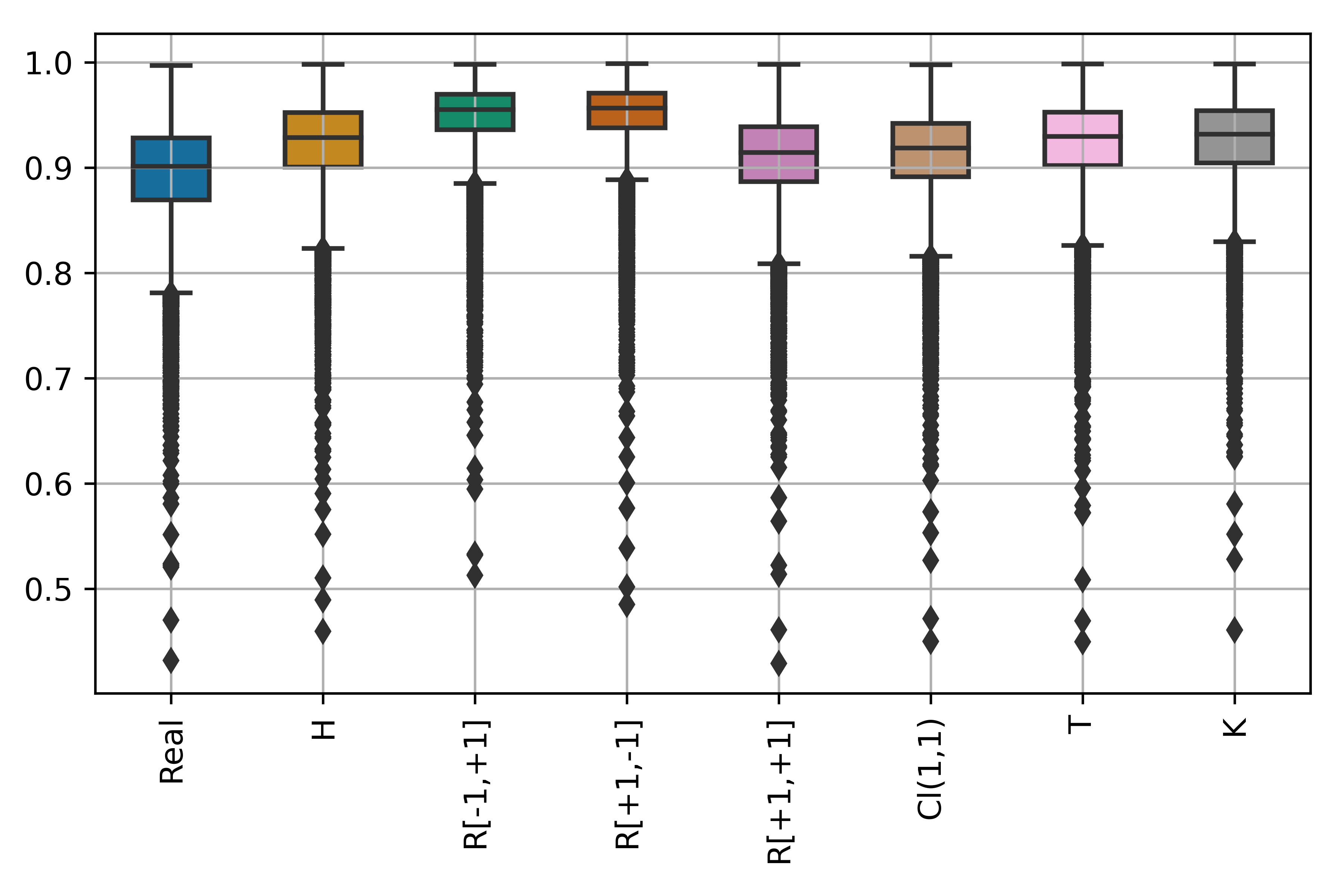}
    % \end{tabular}
    \caption{PSNR and SSIM rates produced by the real and hypercomplex-valued auto-encoders in the test set.}
    \label{fig:boxplot}
\end{figure}

Visually, all auto-encoders performed well in both training and test images, reconstructing the image with a good resemblance. From a quantitative approach, the peak signal-to-noise ratio (PSNR) and structural similarity index (SSIM) were used to evaluate the auto-encoders' performance. These metrics are calculated considering the input image, the desired output, and the auto-encoder's output. On the one hand, the PSNR is a logarithmic scale inversely proportional to the mean squared error, meaning a higher PSNR value indicates a higher quality of reconstruction. On the other hand, the SSIM is a unitary real-valued index, in which higher values represent the high similarity between the input and the output images. Both metrics are robust, and, together, they cover the concept of similarity in great detail. Therefore, this pair of metrics is a good indicator for the model performance. The metrics attained by the ELM auto-encoders are reported in Table \ref{tab:psnr_ssim} as well as the boxplots shown in Fig. \ref{fig:boxplot}.

%To evaluate quantitatively the performance of the ELM auto-encoders two metrics have been used: The peak noise-to-signal ratio (PSNR) and the structural similarity index (SSIM) \cite{wang09ssim}. The PSNR is a logarithmic scale inversely proportional to the mean squared error. Thus, a higher PSNR value means a higher quality of reconstruction. The structural similarity index is a value in the interval $[-1,1]$, in which $1$ represents perfect similarity between  the original and the reconstructed image. We consider the PSNR and SSIM to be fairly different yet robust metrics for the auto-encoding task. The results obtained by the ELM auto-encoders are reported in Table \ref{tab:psnr_ssim} as well as the boxplots shown in Fig. \ref{fig:boxplot}.

\begin{table}[t]
    \centering
        \caption{Average PSNR and SSIM rates achieved by real and hypercomplex-valued auto-encoders.}
    \label{tab:psnr_ssim}
    \renewcommand\arraystretch{1.5}\begin{tabular}{c|cc|cc}
         \multicolumn{1}{c}{} & \multicolumn{2}{c}{\textbf{Train Set}} & \multicolumn{2}{c}{\textbf{Test Set}}  \\ 
         Algebra & PSNR & SSIM & PSNR & SSIM \\ \hline \hline
         Real & ${27.3\pm2.4}$ & ${0.91\pm0.05}$ & ${26.8\pm2.6}$ & ${0.89\pm0.05}$ \\ 
        $\mathbb{H}$ & ${28.9\pm2.5}$ & ${0.93\pm0.04}$ & ${28.5\pm2.7}$ & ${0.92\pm0.05}$ \\ 
        $\R[-1,+1]$ & $\mathbf{31.0\pm2.5}$ & $\mathbf{0.95\pm0.03}$ & $\mathbf{30.5\pm2.7}$ & $\mathbf{0.95\pm0.04}$ \\ 
        $\R[+1,-1]$ & $\mathbf{31.1\pm2.5}$ & $\mathbf{0.95\pm0.03}$ & $\mathbf{30.6\pm2.7}$ & $\mathbf{0.95\pm0.04}$ \\ 
        $\R[+1,+1]$ & ${27.9\pm2.4}$ & ${0.92\pm0.04}$ & ${27.5\pm2.6}$ & ${0.91\pm0.05}$ \\ 
        $C\ell_{1,1}$ & ${28.1\pm2.4}$ & ${0.92\pm0.04}$ & ${27.7\pm2.6}$ & ${0.91\pm0.04}$ \\ 
        $\mathbb{T}$ & ${
        28.9\pm2.5}$ & ${0.93\pm0.04}$ & ${28.5\pm2.7}$ & ${0.92\pm0.05}$ \\ 
        $\K$ & ${28.9\pm2.5}$ & ${0.93\pm0.04}$ & ${28.5\pm2.7}$ & ${0.92\pm0.05}$
        %  Real-valued & $27.3\pm2.4$ & $0.90\pm.05$ 
        %     & $26.8\pm2.7$ & $0.89\pm.05$  \\
        %  $\mathbb{H}$ & $28.9\pm2.5$ & $0.93\pm.04$ 
        %     & $28.5\pm2.7$ & $0.92\pm.05$  \\
        %  $\R[-1,+1]$ & $31.0\pm2.5$ & $\mathbf{0.95\pm.03}$ 
        %     & $30.5\pm2.7$ & $\mathbf{0.95\pm.04}$ \\
        %  $\R[+1,-1]$ & $\mathbf{31.2\pm2.5}$ & $\mathbf{0.95\pm.03}$  
        %     & $\mathbf{30.7\pm2.7}$ & $\mathbf{0.95\pm.04}$ \\
        %  $\R[+1,+1]$ & $30.9\pm2.5 $ & $\mathbf{0.95\pm.03}$ 
        %     & $30.4\pm2.7$ & $\mathbf{0.95\pm.04}$ \\
        % $C\ell_{1,1}$ & ${28.1\pm2.4}$ & ${0.92\pm.05}$  
        %     & ${27.7\pm2.6}$ & ${0.91\pm.05}$ \\
        % $\mathbb{T}$ & ${28.9\pm2.5}$ & ${0.93\pm.04}$  
        %     & ${28.5\pm2.7}$ & ${0.92\pm.05}$ \\
        % $\K$ & ${28.9\pm2.4}$ & ${0.93\pm.04}$  
        %     & ${28.5\pm2.6}$ & ${0.92\pm.04}$
    \end{tabular}
\end{table}

Note, from Table \ref{tab:psnr_ssim}, that the eight auto-encoders yielded similar PSNR and SSIM rates when comparing training and test set results. Thus, the ELMs learned the auto-encoding task with adequate generalization capability. Furthermore, the hypercomplex-valued auto-encoders outperformed the real-valued model by a noticeable margin, as expected for a multidimensional input problem. The auto-encoders based on quaternions, tessarines, Klein four-group, the Cayley-Dickson algebra $\R[+1,+1,]$, and the Clifford algebra $C\ell_{1,1}$ yielded similar performance among the hypercomplex-valued models. In contrast, the ELM models based on the Cayley-Dickson algebras $\R[-1,+1]$ and $\R[+1,-1]$ pulled ahead by a significant margin. They are top-performing models with an almost perfect score in terms of the structural similarity index.

With regards to computational complexity, the hypercomplex-valued ELMs are much more time demanding than their real counterpart. The computational burden is mainly due to the transformations $\Phi_L$ and $\Phi_R$ required on \eqref{eq:CD2Real}, \eqref{eq:CD2Real2}, and \eqref{eq:CayleyLSP-solution}. For example, in our experiment, the training phase is carried out in $3.25$s and $283.76$s for the real and hypercomplex-valued models, respectively. This amounts to a roughly $87$ times faster training step for the real model. However, we would like to highlight that we implemented general-purpose codes for the hypercomplex-valued ELM models. We strongly believe that this gap in time-wise performance can be greatly reduced with the proper implementation of specific hypercomplex-valued operations.

\section{Concluding Remarks} \label{sec:concluding-remarks}

%\textbf{Coloquei o texto que estava aqui no abstract. Reescrever a conclusão fazendo referências as equações, definições e teoremas demostrados no texto. Falta completar as referências!} 

Hypercomplex-valued neural networks are known to perform better than real-valued models for tasks involving high-dimensional inputs \cite{aizenberg11book,minemoto17,Vieira2020ExtremeAuto-Encoding}, such as images, video, and digital signals. In this paper, we present a general framework for hypercomplex-valued extreme learning machines (ELMs). 

Precisely, we first established an equivalence between hypercomplex-valued matrix operations and real-valued linear algebra. We also showed how the solution of a hypercomplex-valued least-squares problem is determined using the well-known real-valued least-squares. Because training an ELM is formulated as a least-squares problem, we provided tools for training general hypercomplex-valued ELM models using real linear algebra operations. Moreover, the real-valued formulation given by \eqref{eq:CayleyLSP-solution} depends almost solely on the multiplication table of the underlying hypercomplex algebra. Because its multiplication table uniquely determines a hypercomplex algebra, the definitions presented in this work allows for the implementation of ELMs in any hypercomplex algebra.

%In Section \ref{subsec:notable_algebras} we provided a handful of examples of hypercomplex algebras with interesting well-known properties. By using these algebras we implemented 7 four dimensional hypercomplex models, which were tested in an auto-encoding task along with a real-valued model of equivalent size. This experiment showed that these models are fairly more adequate for the image processing task when it comes to raw performance metrics, even on networks with comparable number of adjustable parameters. We believe that such an advantage stems from the compact representation of elements in higher dimensional algebras allied to the operations ability to cope with multiple signals at once.

In Section \ref{subsec:notable_algebras}, we provided a handful of examples of hypercomplex algebras with interesting well-known properties. Using these algebras, we implemented seven four-dimensional hypercomplex-valued ELM models besides the traditional real-valued ELM. The neural networks have been used for chaotic time series prediction and an auto-encoding task. On both tasks, the hypercomplex-valued models outperformed the traditional real-valued ELM by a noticeable margin. On the one hand, the ELMs' performance gap is maintained in the time series prediction task throughout a broad range of network architectures. On the other hand, the auto-encoding task shows that the hypercomplex-valued models are fairly more adequate for the image processing when it comes to raw performance metrics. We believe that such an advantage stems from the compact representation of elements in higher dimensional algebras allied to the operation's ability to cope with multiple values at once.

We would like to point out that most works on hypercomplex-valued neural networks focus on well-established well-behaved algebras such as complex and quaternions. However, our experiments suggest that some unexplored algebras, such as the Cayley-Dickson algebras, may perform better in some machine learning tasks. In fact, the algebraic properties (or lack thereof) of the multiplication did not impact the ELM models' performance in some tasks. The auto-encoding experiment, in which commutative algebras such as the tessarines and the Klein four-group have been outperformed by an unusual Cayley-Dickson algebra ($\R[+1,+1]$) with no properties of particular interest, exemplifies this remark. Concluding, unusual hypercomplex algebras with apparently no algebraic properties could be viable alternatives for developing new hypercomplex-valued neural networks and their applications.

\bibliographystyle{IEEEtran}
\bibliography{references.bib}

\end{document}